\documentclass[conference,a4paper]{IEEEtran}
\IEEEoverridecommandlockouts
\usepackage{cite}
\usepackage{amsmath,amssymb,amsfonts}
\usepackage{algorithmic}
\usepackage{graphicx}
\usepackage{textcomp}
\usepackage{xcolor}
\usepackage{url}
\usepackage{multirow}
\usepackage{makecell}
\usepackage{fancyhdr}
\def\BibTeX{{\rm B\kern-.05em{\sc i\kern-.025em b}\kern-.08em
    T\kern-.1667em\lower.7ex\hbox{E}\kern-.125emX}}
\begin{document}

\title{An Efficient FPGA-based Accelerator for Deep Forest\\
\thanks{This work was supported in part by the National Natural Science 
Foundation of China under Grant 62174084, 62104097 and in part by the 
High-Level Personnel Project of Jiangsu Province under Grant JSSCBS20210034, 
the Key Research Plan of Jiangsu Province of China under Grant BE2019003-4.
\textit{(Corresponding author: Zhongfeng Wang.)}}
}

\author{\IEEEauthorblockN{Mingyu Zhu, Jiapeng Luo, Wendong Mao, Zhongfeng Wang}
\IEEEauthorblockA{\textit{School of Electronic Science and Engineering} \\
\textit{Nanjing University}, Nanjing, China \\
Email: mingyu.zhu@smail.nju.edu.cn, luojiapeng1993@gmail.com, wdmao@smail.nju.edu.cn, zfwang@nju.edu.cn}
}

\maketitle

\begin{abstract}
Deep Forest is a prominent machine learning algorithm known for its high
 accuracy in forecasting. Compared with deep neural networks, Deep Forest
 has almost no multiplication operations and has better performance on
 small datasets. However, due to the deep structure and large forest quantity, 
 it suffers from large amounts of calculation and memory consumption. 
 In this paper, an efficient hardware accelerator is proposed for deep 
 forest models, which is also the first work to implement Deep Forest on FPGA. 
 Firstly, a delicate node computing unit (NCU) is designed 
 to improve inference speed. Secondly, based on NCU, an efficient architecture 
 and an adaptive dataflow are proposed, in order to alleviate the problem of 
 node computing imbalance in the classification process. Moreover, an optimized 
 storage scheme in this design also improves hardware utilization and 
 power efficiency. The proposed design is implemented on an FPGA board, 
 Intel Stratix V, and it is evaluated by two typical datasets,
 ADULT and Face Mask Detection. The experimental results show that the
 proposed design can achieve around 40$\times$ speedup compared to that 
 on a 40 cores high performance x86 CPU.
\end{abstract}


\section{Introduction}
With the rapid development of machine learning, deep neural networks (DNN) \cite{2016Deep} 
have achieved great breakthrough in artificial intelligence literature. Though 
DNN has dominated the machine learning research fields nowadays, it has some 
obvious deficiencies such as high computational complexity, slow training speed, 
and lack of flexibility on small datasets. In 2017, a new tree-based ensemble learning 
method, Deep Forest (DF), was proposed by Zhou and Feng \cite{2017Deep}. As shown in 
Fig. 1, its cascade structure makes DF able to do representation learning 
like deep neural networks. As an alternative to 
conventional deep learning methods, it has the following advantages over 
deep neural networks. Firstly, DF has almost no multiplication operations, 
which means low computational complexity. Secondly, DF can perform well 
when there are only small datasets or low-dimension datasets in contrast 
to DNN which requires large datasets. Thirdly, there are less hyperparameters 
in DF than in DNN, which makes DF easy to train. However, as the number of 
forests and the depth of the model increase, the computational complexity grows 
severely. Since the CPU cannot meet the real-time application requirements, it is 
of great necessity to accelerate the inference of the deep forest on hardware. 

\begin{figure}[htbp]
\centerline{\includegraphics[scale=0.39]{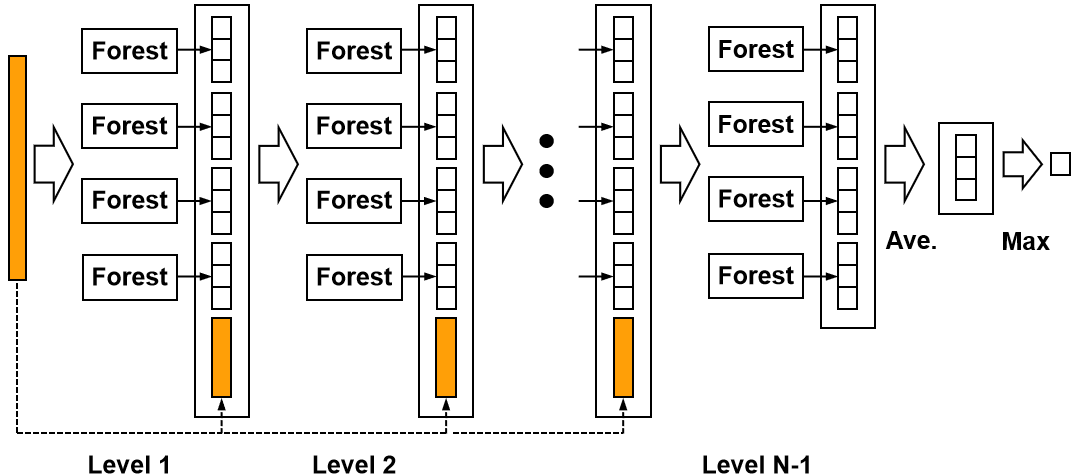}}
\caption{Illustration of the cascade forest structure.}
\label{fig}
\end{figure}

Many hardware accelerators have been developed for tree-based models like Random 
Forest \cite{Breiman2001Random} to improve the speed. When it comes to Deep Forest, 
we face more problems. Firstly, since Deep Forest contains a large-scale ensemble 
of decision trees, it is a big challenge to store all the trees in limited space. 
Secondly, if we traverse all trees in parallel, the problem of node computing 
imbalance will arise due to the different path length of different trees and 
inputs.

In this paper, we propose the first hardware accelerator for DF 
based on FPGA, which improves processing speed with high classification 
accuracy and low power consumption. The main contributions of this paper are 
summarized as follows:
\begin{itemize}
\item A delicate node computing unit (NCU) is designed to decompose the inference 
of a single decision tree into fine-grained logic calculation, in order to 
accelerate the processing. Meanwhile, an optimized storage scheme is introduced 
to store a large number of trees with limited on-chip memory resources.
\item Based on the NCU, a specialized hardware architecture, together with an 
efficient dataflow is proposed to alleviate the problem of node computing 
imbalance in the classification process, while maintaining high classification 
accuracy and low power consumption.
\item The design is implemented on Intel Stratix V FPGA, which is also the first work 
for accelerating Deep Forest on hardware. The experimental results show that the 
proposed design can achieve around 40$\times$ speedup compared to that 
on a 40 cores high performance x86 CPU.
\end{itemize}

\section{Background}

\subsection{Hardware Acceleration for Tree-based Algorithms}

Several previous works have targeted hardware acceleration of the single decision 
tree and Random Forest. In 2012, Van Essen \emph{et al}. \cite{2012Accelerating} 
conducted a comparative study on the acceleration of inference processing of 
random forests by multi-core CPU, GP-GPU and FPGA. The experimental results 
showed that FPGAs can provide the highest performance solution while GP-GPUs 
still have high energy consumption that is sensitive to sample 
size and makes it difficult to be applied to mobile devices or edge devices. 
In their hardware design, 
the calculation cycle of each node is 5 clock cycles which can be further shortened. 
Saqib \emph{et al}. \cite{Saqib2015Pipelined} designed a pipeline structure for DT 
inference, and proposed an acceleration architecture composed of parallel 
processing nodes. Nakahara \emph{et al}. \cite{2017A} proposed a multi-valued 
decision diagrams based on random forests. In the diagram, each 
variable only appears once on the path in order to reduce inference latency. 
The disadvantage is that the number of nodes increases which will slow down the 
training process as a result. Alharam \emph{et al}. \cite{2020Optimized} improved 
the real-time performance of the random forest classifier by reducing the number 
of nodes and branches to be evaluated, and reducing the branch length by numerical splitting.  

However, different from the other tree-based models, Deep Forest is an ensemble 
of ensembles which makes it a big challenge to deal with the large resource 
consumption and the large number of calculations. In addition, the prior works 
mainly focus on shortening the branch length of each tree which brings small 
speed improvement. In this paper, we accelerate the inference of DF with the 
aid of the NCU and propose a special overall architecture for DF based on FPGA.

\subsection{Deep Forest}

The deep forest algorithm includes two parts: Multi-Grained Scanning 
and Cascade Forest. 

Inspired by the layer-by-layer processing of the original features in DNN, Deep 
Forest adopts a cascading structure, as shown in Fig. 1. The cascade forest 
structure stacks multiple forests in this way to obtain enhanced features and 
better learning performance. In the cascade forest, the input of the first layer 
is the feature vector of the instance, and the output of each layer is a set of 
class vectors. The output vectors of the previous layer and the original feature 
vector are concatenated together as the input of the next layer. Here we use two 
random forests \cite{Breiman2001Random} and two completely-random tree forests 
\cite{2008Spectrum} in each layer.

Fig. 2 illustrates the generation of the class vector. The traversed paths of the 
instance are highlighted in orange. For each instance, each forest averages the 
percentages of different classes of training data given by all trees in the same 
forest.

\begin{figure}[htbp]
\centerline{\includegraphics[scale=0.29]{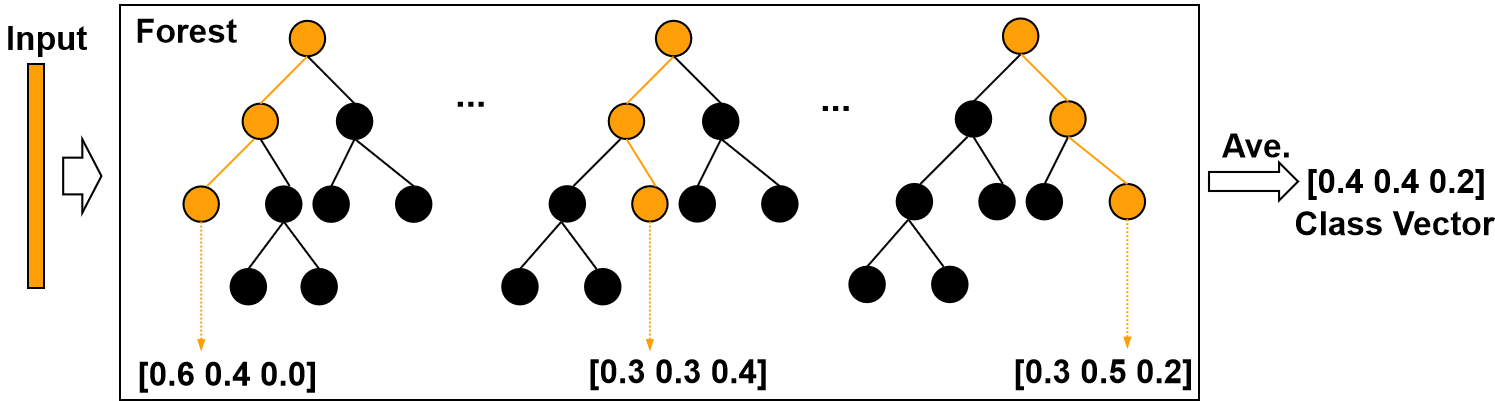}}
\caption{Illustration of class vector generation.}
\label{fig}
\end{figure}

The overall procedure of Deep Forest uses the 
multi-grained scanning process to enhance the cascade forest. By using multiple 
sizes of sliding windows, the transformed feature vectors contain more 
information and different kinds of outputs are sent to the corresponding layer 
of the cascade forest. DF terminates training when the performance cannot be improved.

\section{Proposed design}

\subsection{Node Computing Unit}\label{AA}

Deep Forest contains a cascade structure of ensemble trees, which makes it have 
higher computational complexity than other tree-based models, so it is important 
to reduce memory requirements and improve inference speed. Firstly, the storage 
scheme optimizes the format used to store the trees, while including all the 
information of each node in a 32-bit word. Secondly, we propose a 
computation-efficient node computing unit (NCU) and it can shorten the node 
operation period to 4 clock cycles while \cite{2012Accelerating} uses 
5 clock cycles. 

\begin{figure}[htbp]
\centerline{\includegraphics[scale=0.37]{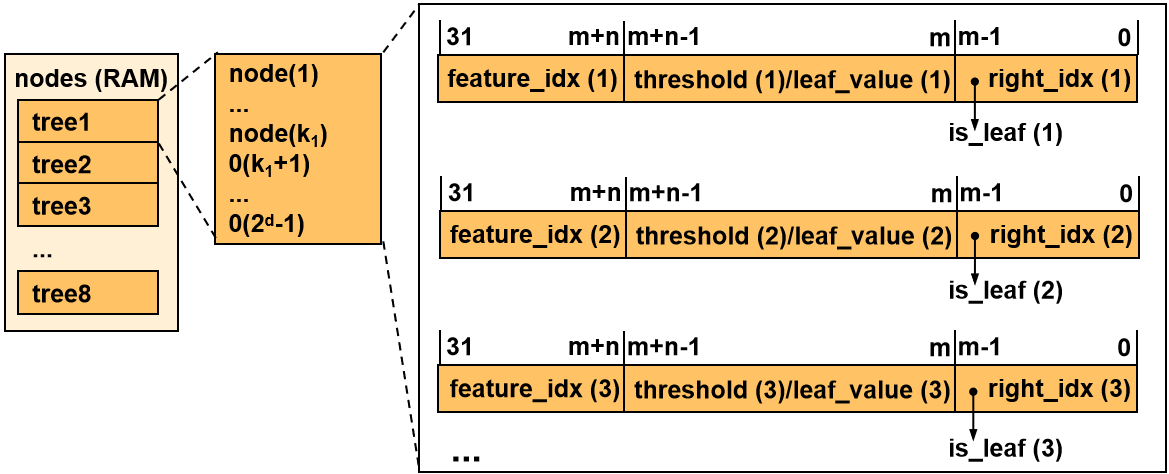}}
\caption{Storage of trees and memory layout of each node.}
\label{fig}
\end{figure}

In traditional storage scheme of tree-based models, the information of a single 
tree includes address of feature, threshold and addresses of the left and right 
child nodes. Our goal is to store each node of a tree in a 32-bit word. 
However, if the address of feature occupies 8 bits and the threshold 
occupies 16 bits, the left 8 bits memory is not enough for 
the addresses of all child nodes of an 8-depth tree. To tackle the 
problem, we propose an optimized storage scheme. Fig. 3 shows the storage of 
trees and the memory layout of each node. In our design, one nodes RAM stores 
the information of 8 trees of maximum depth \emph{d}, and each tree has 
at most \emph{$2^{d}$}-1 nodes. The format of each node includes 
three fields. For non-leaf nodes, the first field stores the \emph{feature\_idx} 
deciding which feature will be used. The second field stores the \emph{threshold} 
(\emph{n} bits) which will be compared with the selected feature. For the 
addresses of child nodes, we use the pre-order traversal method to store the 
nodes, which means the memory position of a left child node always follows its 
parent node. In this way, we can deduce the address of a left child node from 
its parent node address. Therefore, the third field only stores the address of 
the right child (\emph{m} bits) with a sign bit. When it comes to the leaf nodes, 
the sign bit of the \emph{right\_idx} turns to 1, which distinguishes 
two kinds of nodes. For the leaf nodes, the second field stores the 
\emph{leaf\_value} (\emph{n} bits) which is the output of the tree. In our design, 
\emph{m} is 9, and without the address of the left child node, 
20\% storage space of trees are saved.

\begin{figure}[htbp]
\centerline{\includegraphics[scale=0.38]{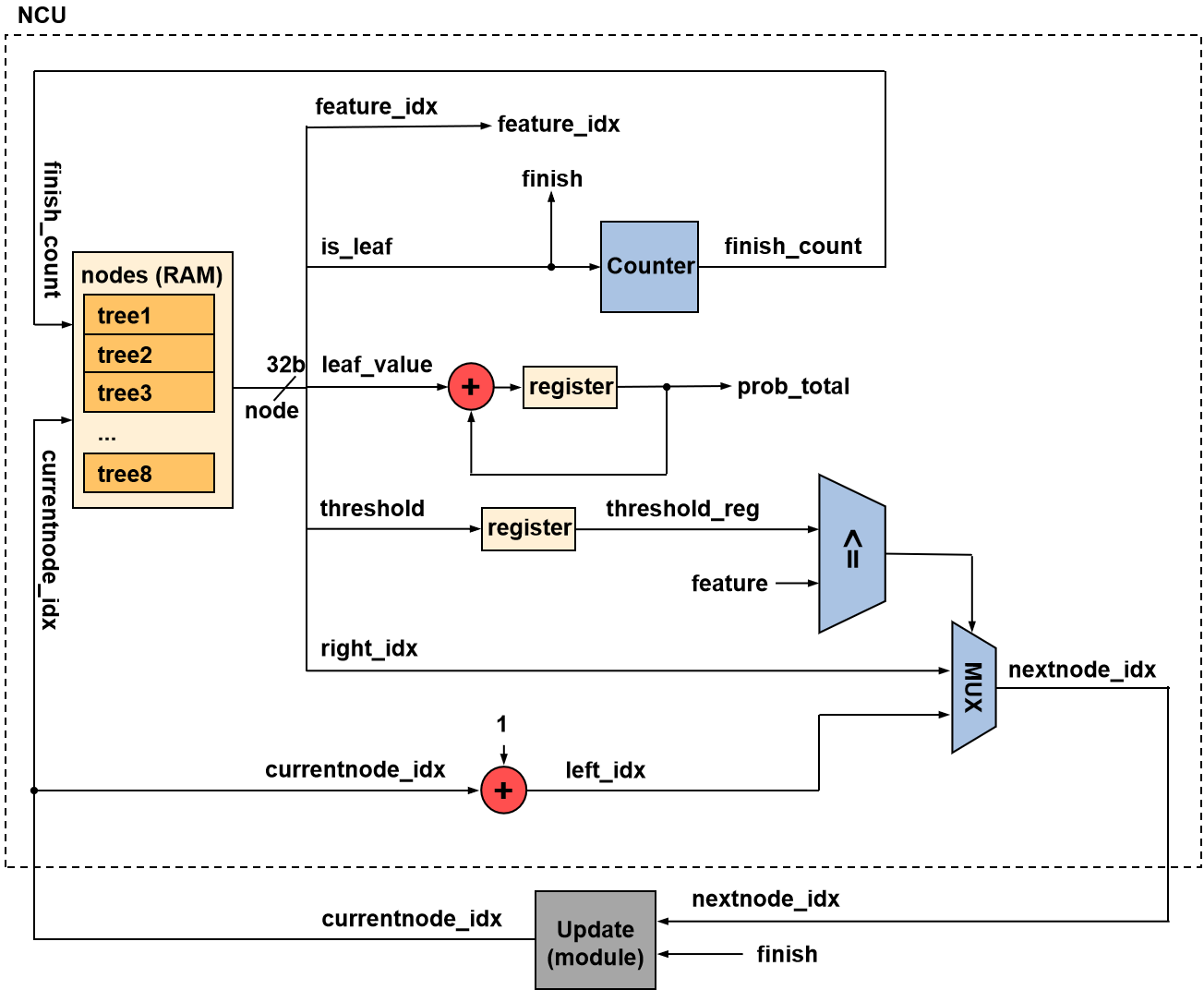}}
\caption{The NCU and the node updating module. 
The NCU includes the nodes RAM memory storing the information of all trees in a group 
and the logic carrying out the comparison and accumulating the output of the trees.}
\label{fig}
\end{figure}

In our design, all trees in a forest are divided into several groups and 
one NCU takes charge of a group of trees stored in one RAM.
The description of the NCU and the node updating module is shown in Fig. 4. The 
design includes the nodes RAM memory, a main input, \emph{currentnode\_idx}, 
which is used to store the address of current node, a main output, 
\emph{nextnode\_idx}, which is used to store the address of next node, and the 
logic that carries out the comparison and accumulates the \emph{prob\_total}, 
which is the output of the trees. To start, as the nodes RAM stores 8 
trees, the \emph{currentnode\_idx} is concatenated with the \emph{finish\_count} 
to get the information of the current node from the nodes RAM. Then, the 
\emph{feature\_idx} of the non-leaf nodes is concatenated with the 
\emph{finish\_count} to select one feature from all input features. After a clock 
cycle, we get the \emph{feature} and it is compared with the \emph{threshold\_reg}. 
In our design, we use the combinational logic instead of the sequential logic to 
implement the \textbf{comparator}. Finally, the comparison result is sent to a 
\textbf{multiplexer}, deciding whether the left child or the right child will be 
the next node. To obtain the address of the left child, \emph{left\_idx}, we add 
1 to the \emph{currentnode\_idx}. As the output, the \emph{nextnode\_idx} 
needs to be sent to the \textbf{update} module to get the \emph{currentnode\_idx} 
which will be used in the next round.

If the current node is a leaf node, the \emph{is\_leaf} value is 1 and 
is sent to the \textbf{counter} to get the counting result, \emph{finish\_count}. 
Meanwhile, the \emph{leaf\_value} is accumulated with the previous 
\emph{prob\_total} to get a new one. As a result, the overall calculation cycle 
of each node is shortened to four clock cycles which is one clock cycle less 
than that of \cite{2012Accelerating}. 

\subsection{Overall Architecture and Dataflow}
For Deep Forest inference, the cascade forest occupies most of the time, so it’s 
crucial to accelerate this part on hardware. We insert a pipeline at the end of 
each layer in order to accelerate the processing. The proposed overall 
architecture is illustrated as Fig. 5. In our design, each layer occupies 
different on-chip resource. There are two forests in one layer, each of which 
is processed by one \textbf{PE}. A forest consists of 32 trees and 
8 trees are packed into a group and are processed by one and the same 
\textbf{NCU}. Therefore, each \textbf{PE} is composed of 8 \textbf{NCU}s 
and all of them are run in parallel. The final prediction is obtained by averaging 
the output of the last layer, and then sent to the off-chip \textbf{DRAM}.

There are three kinds of buffers to store data on the chip. 
\textbf{Input Buffer}, whose basic unit is RAM, stores three feature vectors 
produced by the multi-grained scanning, supposing we use three sizes of sliding 
windows. \textbf{Layer 1$\sim$4 Buffer} and \textbf{Output Buffer} 
store the input features of layer 1$\sim$4 and the output vector of 
the whole on-chip logic.

\textbf{Average} is composed of adders and a shift register. The adders accumulate 
the classification results of all NCUs in one PE. As there are 32 trees 
in a forest, a shift register is used to get the mean value of all trees. When 
the averaging is finished, the result will be stored in a register.

\textbf{Update} contains a counter and a register. The counter records the period 
of the NCU. Once the period reaches four clock cycles, the address of the current 
node in a register, \emph{currentnode\_idx}, is replaced by the address of the 
next node, \emph{nextnode\_idx}. When the module receives the finish signal from 
the corresponding NCU, \emph{currentnode\_idx} turns to zero.

\textbf{Controller} receives the finish signals of all layers and counts the 
number of final results. It takes charge of data transport from off-chip DRAM 
to Input Buffer and from registers to Layer 1$\sim$4 Buffer. It is worth 
noting that we concatenate the data fetched from Layer 1$\sim$4 Buffer with the 
original feature vector fetched from the corresponding Input SRAM when Layer 
1$\sim$4 Buffer receives the signal from the controller.  

\begin{figure*}[htbp]
\centerline{\includegraphics[scale=0.30]{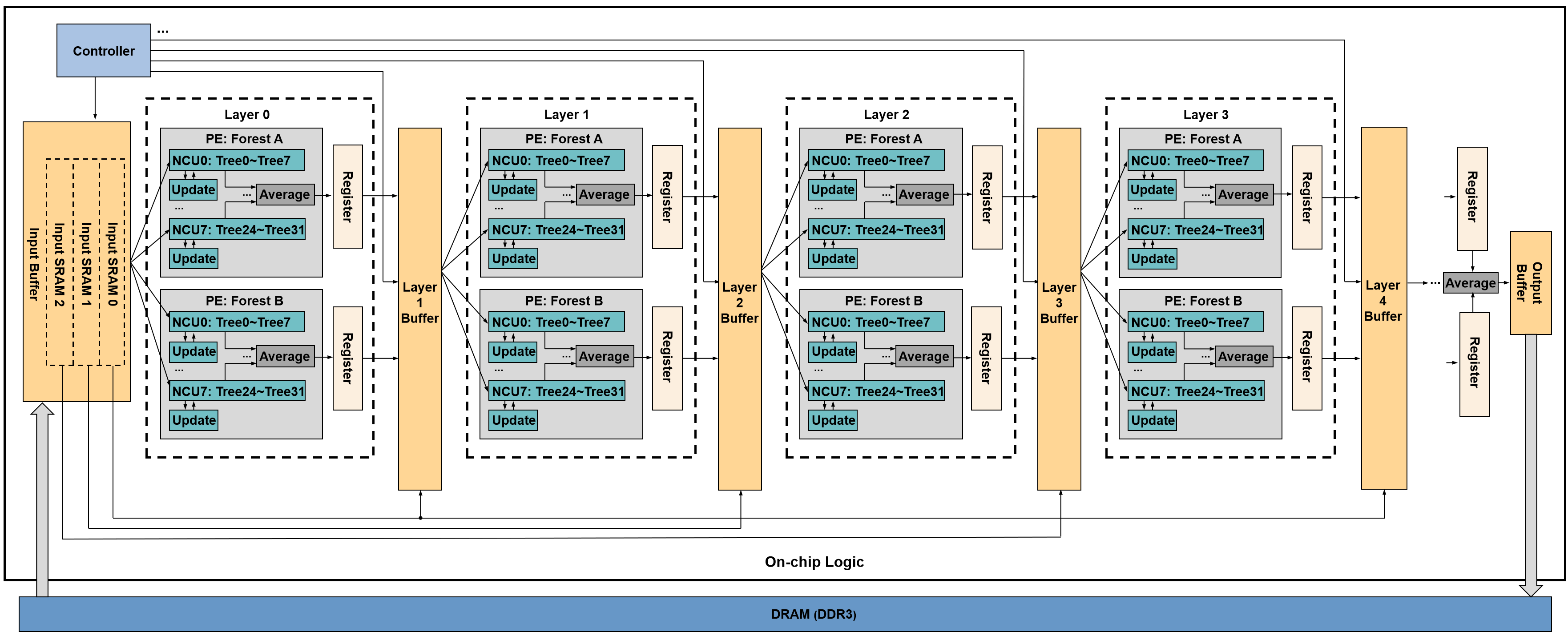}}
\caption{The overall architecture. Each layer contains two PEs and each PE contains eight NCUs, each of which processes eight trees.}
\label{fig}
\end{figure*}

Since different samples have different path lengths, it will cause the problem 
of node computing imbalance. To solve this problem, we propose the following 
dataflow. All NCUs in one PE are run in parallel, and each NCU is responsible for 
a group of trees instead of only one decision tree as the traditional methods. 
Once the NCU finishes a tree, it immediately processes the next one. The 
decision trees in the same group are sequentially traversed in a serial manner. 
In this way, we can mitigate the impact of gaps between various path lengths. 
Moreover, we insert a pipeline at the end of each layer to improve data throughput.

\section{Experiments}

\subsection{Configuration}
In this section, we use two DF models trained on ADULT \cite{2013UCI} and Face Mask 
Detection \cite{TP-toolbox-web} respectively. Face Mask Detection is a new image 
dataset distinguishing whether a person wears a mask correctly or not. For ADULT, 
the multi-grained scanning is abandoned considering that the features have few 
sequential or spacial relationships. There are 4 layers in this model 
and each layer consists of one completely-random tree forest and one random 
forest, each containing 32 trees. For Face Mask Detection, 3 
sizes of sliding windows are used in the multi-grained scanning. The cascade 
forest is composed of 3 layers and the configuration of each layer is 
the same as the model trained on ADULT.

\subsection{Results}

We run the above two models on Intel Xeon Gold 6148 CPU (40 
cores), and our hardware design is implemented on FPGA (Intel Stratix V), 
reaching a clock frequency of 400 MHz.

The proposed design decreases the usage of on-chip resourses. The resource 
utilization of our design on Intel Stratix V is shown in Table I. Because of 
the different data sizes, the models trained by the two datasets are implemented 
on different chips.

\begin{table}[htbp]
    \centering
    \caption{The Resource Utilization of Our Design}
    \begin{tabular}{|c|c|c|}
    \hline
    \textbf{}&\multicolumn{1}{|c|}{\textbf{ADULT}}&\multicolumn{1}{|c|}{\textbf{Face Mask Detection}} \\
    \cline{2-3} 
    \hline
    Device& Stratix V 5SGXMA3& Stratix V 5SGXEAB\\
    \hline
    ALMs& 41,377 / 128,300 (32\%) & 213,104 / 359,200 (59\%)\\
    \hline
    Memory (KB)& 314 / 2,392.5 (13\%)& 420 / 6,600 (6\%)\\
    \hline
    DSP Blocks& 0& 0\\
    \hline
    \end{tabular}
    \label{tab1}
\end{table}

Table II shows the comparison of our implementation on FPGA at a clock frequency of 
400 MHz with CPU. We evaluate the throughput rate on the two datasets, 
and find that our design achieves great speedup compared to the 40 cores 
high performance x86 CPU. It increases the throughput rate 40 
times on ADULT and 1,871 times on Face Mask Detection. The proposed 
design also brings a great improvement on the latency.

\begin{table}[htbp]
    \centering
    \caption{The Comparison of Our Design on FPGA (400MHz) with CPU}
        \begin{tabular}{|c|c|c|c|c|c|}
        \hline
        \textbf{}&\multicolumn{3}{|c|}{\shortstack{\textbf{Throughput Rate} \\\textbf{(Ksamples/s)}}}&\multicolumn{2}{|c|}{\shortstack{\textbf{Latency} \\\textbf{($\mu$s)}}} \\
        \cline{2-6} 
        \textbf{} & \textbf{\textit{CPU}}& \textbf{\textit{Ours}}& \textbf{\textit{Speedup}}& \textbf{\textit{CPU}}& \textbf{\textit{Ours}} \\
        \hline
        ADULT& 37.59& 1,525&40$\times$ & 34,000& 2.52\\
        \hline
        Face Mask Detection& 0.75& 1,413&1,871$\times$ & 877,000& 2.36\\
        \hline
        \end{tabular}
        \label{tab1}
\end{table}

\begin{table}[htbp]
    \centering
    \caption{The Comparison of Our Work with \cite{2012Accelerating}}
    \begin{tabular}{|c|c|c|}
    \hline
    \textbf{}&\multicolumn{1}{|c|}{\textbf{Ours}}&\multicolumn{1}{|c|}{\textbf{\cite{2012Accelerating}}}\\
    \cline{2-3} 
    \hline
    Platform& \makecell[c]{Intel Stratix V  \\5SGXMA3} & \makecell[c]{Xilinx Virtex 6 \\XC6VLX} \\
    \hline
    Number of FPGAs& 1& 2 \\
    \hline
    Frequency (MHz)& 400 & 100 \\
    \hline
    Throughput Rate (Ksamples/s)& 1,525 & 31,250 \\
    \hline
    Power (W)& 2.64 & 11 \\
    \hline
    Energy Efficiency (GOPS/W)& 517,117 & 499,968 \\
    \hline
    \end{tabular}
    \label{tab1}
\end{table}

Since the complexity of the deep forest algorithm is higher than the other 
tree-based algorithms, the energy efficiency becomes another important performance 
when these methods are implemented on FPGA. Table III shows the comparison of our 
work with \cite{2012Accelerating}. In our design, the energy efficiency on ADULT 
surpasses that of \cite{2012Accelerating}, but the latter needs more than one FPGA 
to implement the same number of trees as one layer in our model.

\section{Conclusion}

In this paper we propose an efficient hardware architecture for the deep forest 
model which is also the first work to accelerate DF. Implemented on Intel Stratix 
V FPGA, the proposed design achieves at least 40$\times$ speedup compared 
to that on a 40 cores high performance x86 CPU. Since there are no 
previous works on hardware acceleration of DF, we compare it with the hardware 
accelerator of Random Forest and find our design has comparable energy 
efficient while consuming less hardware resources. There are many potential 
applications for the proposed design, especially some classification tasks 
on mobile devices.

\bibliographystyle{unsrt}
\bibliography{myref}

\begin{thebibliography}{10}

\bibitem{2016Deep}
Ian Goodfellow, Yoshua Bengio, and Aaron Courville.
\newblock {\em Deep Learning}.
\newblock Deep Learning, 2016.

\bibitem{2017Deep}
Z.~H. Zhou and J.~Feng.
\newblock Deep {F}orest: {Towards An Alternative to Deep Neural Networks}.
\newblock 2017.

\bibitem{Breiman2001Random}
Breiman.
\newblock Random forests.
\newblock {\em {Machine Learing}}, 2001,45(1)(-):5--32, 2001.

\bibitem{2012Accelerating}
Brian Van~Essen, Chris Macaraeg, Maya Gokhale, and Ryan Prenger.
\newblock {Accelerating a Random Forest Classifier: Multi-Core, GP-GPU, or
  FPGA?}
\newblock In {\em IEEE International Symposium on Field-programmable Custom
  Computing Machines}, 2012.

\bibitem{Saqib2015Pipelined}
Saqib, Dutta, Plusquellic, Ortiz, Pattichis, and MS.
\newblock {Pipelined Decision Tree Classification Accelerator Implementation in
  FPGA (DT-CAIF)}.
\newblock {\em {IEEE Transactions On Computers}}, 2015,64(1)(-):280--285, 2015.

\bibitem{2017A}
H.~Nakahara, A.~Jinguji, S.~Sato, and T.~Sasao.
\newblock {A Random Forest Using a Multi-valued Decision Diagram on an FPGA}.
\newblock In {\em 2017 IEEE 47th International Symposium on Multiple-Valued
  Logic (ISMVL)}, 2017.

\bibitem{2020Optimized}
A.~K. Alharam and A.~Shoufan.
\newblock {Optimized Random Forest Classifier for Drone Pilot Identification}.
\newblock In {\em 2020 IEEE International Symposium on Circuits and Systems
  (ISCAS)}, 2020.

\bibitem{2008Spectrum}
Fei~Tony Liu, Kai~Ming Ting, Yang Yu, and Zhi~Hua Zhou.
\newblock {Spectrum of Variable-Random Trees}.
\newblock {\em Journal of Artificial Intelligence Research}, 32(1):355--384,
  2008.

\bibitem{2013UCI}
K.~Bache and M.~Lichman.
\newblock {UCI Machine Learning Repository}.
\newblock 2013.

\bibitem{TP-toolbox-web}
Péter Baranyi.
\newblock {TP} {T}oolbox.
\newblock
  \url{https://www.kaggle.com/ashishjangra27/face-mask-12k-images-dataset}.

\end{thebibliography}

\vspace{12pt}
\color{red}
\end{document}